\pdfoutput=1

\documentclass[11pt]{article}

\usepackage{acl}

\usepackage{times}
\usepackage{latexsym}

\usepackage[T1]{fontenc}

\usepackage[utf8]{inputenc}

\usepackage{microtype}

\usepackage{todonotes,comment,pbox,multirow,arabtex,utf8,pifont,graphics,amsmath,textcomp,subcaption,array,float,adjustbox,subcaption,caption,pbox,amssymb,amsmath,booktabs}
\usepackage{algorithm}
\usepackage{algorithmic}
\usepackage{tabularx}
\usepackage{colortbl}
\usepackage{xifthen}

\usepackage{nicefrac,natbib}
\usepackage{caption,subcaption}
\usepackage{pgfplots,pgfplotstable}
\pgfplotsset{compat=1.8}
\usepgfplotslibrary{statistics}

\newcommand\lexDict{{\sc Lex$_{Dict}$}}
\newcommand\lexRand{{\sc Lex$_{Rand}$}}
\newcommand\LexPred{{\sc Lex$_{Pred}$}}

\newcommand\ecRand{{\sc EC$_{Rand}$}}
\newcommand\ecSPF{{\sc EC$_{SPF}$}}
\newcommand\mlRand{{\sc ML$_{Rand}$}}
\newcommand\mlSPF{{\sc ML$_{SPF}$}}
\newcommand\BT{{\sc BT}}

\title{Data Augmentation Techniques for Machine Translation\\ of Code-Switched Texts: A Comparative Study}

\author{Injy Hamed,$^{1,2}$ Nizar Habash,$^{1}$ Ngoc Thang Vu$^{2}$ \\
  $^1$Computational Approaches to Modeling Language Lab, 
  New York University Abu Dhabi \\
  $^2$Institute for Natural Language Processing, University of Stuttgart\\
  \texttt{\{injy.hamed,nizar.habash\}@nyu.edu} \\\texttt{thang.vu@ims.uni-stuttgart.de}
  }

\begin{document}
\maketitle
\begin{abstract}
Code-switching (CSW) text generation has been receiving increasing attention as a solution to address data scarcity. In light of this growing interest, we need more comprehensive studies comparing  different augmentation approaches. In this work, we compare three popular approaches: lexical replacements, linguistic theories, and back-translation (BT), in the context of Egyptian Arabic-English CSW. We assess the effectiveness of the approaches on machine translation and the quality of augmentations through human evaluation. We show that BT and CSW predictive-based lexical replacement, being trained on CSW parallel data, perform best on both tasks. Linguistic theories and random lexical replacement prove to be effective in the lack of CSW parallel data, where both approaches achieve similar results. 

\end{list} 
\end{abstract}

\setcode{utf8}

\section{Introduction}
Code-switching (CSW) is the alternation of language in text or speech, which can occur across different levels of granularity: sentences, words and morphemes. CSW is a common phenomenon in Arabic-speaking countries, as in other multilingual communities. Given that Arabic is a morphologically rich language \cite{HEH12}, speakers produce morphological CSW, as illustrated below:
\begin{center}
\<حالاً>
algorithm+\<الـ>
implement+\<هـ>
\<طيب>
 $\Leftarrow$
\\
`\textit{Okay}, \textit{I'll}+implement \textit{the}+algorithm \textit{right away}'
\end{center}
CSW introduces a set of challenges to NLP systems, not least of which is data scarcity. 
This is attributed to CSW being a predominantly spoken phenomenon, only recently increasing in written form on social media. Data augmentation has proved to be a successful workaround for this limitation.
Researchers have investigated several techniques for CSW data augmentation, including learning CSW points \cite{SL08,GVS21}, 
lexical replacements \cite{AGE+21,XY21,GVS21,hamed2022investigating}, linguistic theories \cite{PBC+18,LYL19,hussein2023textual}, neural-based approaches \cite{CCL18,WMW+18,WMW+19,MLJ+19,SZY+19,LV20}, and machine translation (MT) \cite{VLW+12,TKJ21}. With increasing efforts in this area, we need more comparative studies to better understand the merits and requirements of different approaches. 

Efforts along these lines include the work of \newcite{PC21}, where different linguistic-driven and lexical replacement techniques were compared through human evaluation, but not for NLP tasks. \newcite{WMW+18} propose the use of pointer-generator network and compare it against the equivalence constraint (EC) theory \cite{Pop00} and random lexical replacement for LM, without human evaluation. 
\newcite{hamed2022investigating} compare multiple lexical replacement techniques covering human evaluation and performance on language modeling (LM), automatic speech recognition (ASR), MT, and speech translation. \newcite{hussein2023textual} compare using the EC theory and random lexical replacement for LM and ASR, also reporting human assessments. 

In this work, we compare three main approaches:
\textbf{lexical replacements}, \textbf{linguistic theories}, and \textbf{back-translation (BT)}. 
We evaluate the approaches for both naturalness of CSW generations and 
performance on MT, where we focus on CSW Egyptian Arabic-English to English translation. 
The rationale for our focus on MT is the scarcity of work around data augmentation 
as opposed to LM and ASR. Furthermore, previous work on MT focuses on lexical replacements \cite{MLJ+19,SZY+19,AGE+21,XY21,GVS21,hamed2022investigating} and BT \cite{TKJ21}, without substantial comparison between approaches. 
Through our comparative study, we provide answers to the following research questions:
\begin{itemize}
    \item \textbf{RQ1:} Which augmentation technique perform best in zero-shot and non-zero-shot settings (with/without the availability of CSW parallel corpora) for MT?
    \item \textbf{RQ2:} Does generating more natural synthetic CSW sentences entail improvements in MT?
\end{itemize}
\section{Data Augmentation Techniques}
\begin{figure}[t]
\centering
\subfloat{
  \label{fig:aug_example_fig}
  \includegraphics[width=0.37\textwidth]{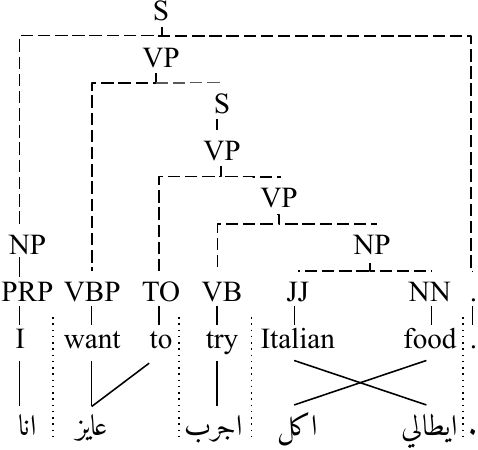}}
\vspace{0.4cm}
\\
\subfloat{%
\small
      \begin{tabular}[h]{|l|r|}
      \hline
         Approach& \multicolumn{1}{c|}{Augmentation Example}\\
         \hline\hline
         \lexDict & \<اجرب اكل ايطالي .> wanting \<انا> $\Leftarrow$\\
         \lexRand & . \< اجرب اكل ايطالي > want to \<انا> $\Leftarrow$\\ 
         \LexPred & . Italian \<انا عايز اجرب اكل> $\Leftarrow$\\\hline\hline
         {\sc EC} & . \< اكل ايطالي > try \<انا عايز> $\Leftarrow$\\
         {\sc EC} \& {\sc ML} & . Italian food \<انا عايز اجرب> $\Leftarrow$\\\hline
      \end{tabular}
      \label{table:aug_example_table}}
\caption{An example showing possible augmentations by the different techniques. We show the parse tree for the English sentence and word alignments. The permissible switching points under the EC theory are shown by the dotted lines.}
\label{fig:aug_example}
\end{figure}

We provide an overview on the investigated techniques.\footnote{We make our  relevant code available at:\\\url{http://arzen.camel-lab.com/} \label{footnoteXcode}} 
Our aim is to augment Arabic-to-English parallel sentences, 
converting the source side of the parallel data from monolingual Arabic to CSW Arabic-English, further extending the MT training data with CSW instances. In Figure \ref{fig:aug_example}, we provide an example showing possible augmentations 
across techniques. More examples are shown in Table~\ref{table:aug_examples}. 
\subsection{Lexical Replacements}
We investigate the following three approaches:
\paragraph{Dictionary Replacement:} 
We replace ${x}$ random Arabic words on the source side with English gloss entries. We obtain the gloss entries using MADAMIRA \cite{PAD+14}, an Arabic morphological analyzer and tagger. Such a specialized analysis system is required for this task as Arabic is morphologically rich and orthographically ambiguous. We refer to this approach as \lexDict.

\paragraph{Aligned with Random CSW Point Assignment:} 
We augment the Arabic-to-English parallel sentences by randomly picking 
${x}$ source-target aligned words (using intersection alignments) and replacing the source words with their counterpart words on the target side. In \newcite{hamed2022investigating}, the authors investigated two types of alignments for performing source-target replacements: (1)~word replacements using intersection alignments and (2)~segment replacements where grow-diag-final alignments are used to identify aligned segments. Given that segment replacements were shown to be superior, we follow that setup in our experiments. We refer to this approach as \lexRand. 

\paragraph{Aligned with Learnt CSW Point Prediction:} 
Similar to the previous approach, we perform target-to-source replacements; however, the choice of words on the target side to be inserted into the source side is based on a CSW predictive model \cite{AGE+21,hamed2022investigating}. The model is trained to identify words on the target side that would be plausible CSW words on the source side. The task of CSW point prediction is modeled as a sequence-to-sequence classification task. The neural network takes as input the target sentence word sequence $x=\lbrace x_1, x_2, .. , x_N \rbrace$, where ${N}$ is the length of the sentence. The network outputs a sequence $y=\lbrace y_1, y_2, .. , y_N \rbrace$, where $y_n\in \lbrace$1,0$\rbrace$ represents whether the word $x_n$ is a plausible CSW word or not. 
To obtain the training data for the predictive model, we utilize a limited amount of CSW Egyptian Arabic-English to English parallel sentences, where we tag the words on the target side as 0 or 1 based on whether they appear as CSW words on the source side or not. This is done using a matching algorithm described in \newcite{hamed2022investigating}. 
The CSW predictive model is then trained by fine-tuning mBERT on this data.\footnote{The hyperparameters are shown in Appendix \ref{sec:appendix-MThyperparameters}.} Afterwards, to augment Arabic-to-English parallel data, we use the model to identify CSW candidates on the target side which are inserted in the source side using segment replacements. For a detailed description of this approach, see \newcite{hamed2022investigating}. We refer to this approach as \LexPred. 

\subsection{Linguistic Theories}
We cover the following two linguistic theories:

\paragraph{Equivalence Constraint (EC) Theory:}
The EC Theory \cite{Pop00} is an alternational model for CSW, where there are no defined matrix and embedded languages. 
Instead, the theory states that code-switching can occur at points where the surface structures of both languages map onto each other. 
In the example in Figure \ref{fig:aug_example}, the permissible alternations are indicated by dotted lines. 
Generating ``\<أكل> Italian'' and ``Italian \<أكل>'' is not allowed as the syntactic rules of both languages are different (Arabic adjectives follow the nouns they modify). 

\paragraph{Matrix Language Frame (MLF) Theory:} 
The MLF Theory \cite{myers1997duelling}, on the other hand, 
is an insertional model. It is based on the identification of a matrix language, to which constituents of the embedded language are inserted such that the sentence follows the grammatical structure of the matrix language, and the embedded language is inserted at grammatically correct points.  Unlike the EC theory, replacements from the embedded to matrix language are not allowed within nesting sub-trees. Replacements of closed-class constituents are also not allowed, including determiners, quantifiers, prepositions, possessive, auxiliaries, tense morphemes, and helping verbs. 

For both linguistic theories, we use the GCM tool \cite{RSG+21}.\footnote{The tool takes $\approx$ 12 hours to augment 309$k$ parallel sentences for each linguistic theory.} 
The tool provides multiple augmentations per source-target parallel sentence, following a linguistic theory. To sample from these generations, it provides two sampling approaches: random and Switch-Point Fraction (SPF) \cite{PBC+18}. 
In random sampling, $k$ generations are picked randomly. In SPF sampling, the generations are ranked based on their SPF distribution compared to a reference distribution  obtained from real CSW data and the top-$k$ generations are chosen. SPF is calculated as the number of switch points divided by the total number of language-dependent tokens in a sentence.\footnote{The current version of the GCM tool provides an implementation for Switch Point (SP) which does not account for the number of tokens in the sentence. We implement our own code for ranking based on SPF. See footnote \ref{footnoteXcode}.} We set $k$ to 1, which is unified across all techniques. We include both sampling approaches, where we refer to the variants as \ecRand, \ecSPF, \mlRand, and \mlSPF.

\begin{table}[]
\centering
\setlength{\tabcolsep}{2pt}
\begin{tabular}{|l|l|c|r|}
\hline
\multicolumn{2}{|c|}{\textbf{Model}} & \multicolumn{1}{c|}{\textbf{top-\textit{k}}} & \multicolumn{1}{c|}{\textbf{\#Aug}} \\\hline
1&[en-csw\&ar] & 1  & 0.1$k$ \\\hline
2&[en-csw\&ar]$\rightarrow$[en-csw] & 1  & 10$k$ \\\hline 
3&[en-csw\&ar]+[en-en]$\rightarrow$[en-csw] & 1 & 19$k$\\\hline
4&[en-csw\&ar]+[en-en]$\rightarrow$[en-csw] & 19 & 151$k$ \\\hline
\end{tabular}
\caption{The number of CSW generations (\#Aug) obtained from the different BT setups: (1) BT model trained on English to Arabic and English to CSW Arabic-English parallel sentences, (2) same as 1 and followed by fine-tuning using the English to CSW Arabic-English parallel data, (3) similar to 2, with appending English sentences to both sides of the training data, and (4) same as 3 with utilizing the top-19 hypotheses.
}
\label{table:BT_CM_aug_figures}
\end{table}

\subsection{Back-translation}
Despite BT \cite{sennrich2015improving} being a well-known data augmentation technique, it has received little attention in the scope of CSW \cite{TKJ21}. 
In this approach, we train a BT model to translate English sentences to CSW Arabic-English. We then use this model to translate the target side of the Arabic-to-English parallel sentences, generating synthetic CSW Arabic-English to English parallel sentences. The BT model is trained on a limited amount of English to CSW Arabic-English parallel sentences and a larger amount of English to Arabic parallel data. However, when using this model to translate 309$k$ English sentences, only 109 CSW sentences are generated, with the rest of the translations being monolingual Arabic. This is due to the training data of the BT model only constituting of 0.7\% of sentences having CSW. 
We boost the number of generated CSW synthetic sentences through the following steps:
\begin{enumerate}
    \item We fine-tune the model using the English to CSW Arabic-English parallel data.
    \item In the BT model training data, we further append the English sentences in the parallel corpus to both source and target sides. 
    \item At inference, instead of obtaining the top-1 hypothesis for each English sentences, we utilize the top-$k$ hypotheses and obtain the CSW translation with the highest confidence score. We set $k$ to 19, where we could not further increase the value of ${k}$ due to computational constraints.
\end{enumerate}
In Table \ref{table:BT_CM_aug_figures}, we show the effect of each step on the number of obtained CSW generations, reaching a total of 151$k$ CSW augmentations by applying all three steps (augmenting 49\% of original sentences). 

\section{Experimental Setup}
\label{sec:experimental_setup}
\subsection{Data}
\label{sec:data}
We use two sources of data: (1) ArzEn-ST \citep{hamed2022arzenST}, which is a CSW-focused parallel corpus and (2) monolingual Egyptian Arabic-to-English parallel corpora. 
ArzEn-ST contains English translations of a CSW Egyptian Arabic-English speech corpus \cite{HVA20} gathered through informal interviews with bilingual speakers. The corpus is divided into train, dev, and test sets having $3.3k$, $1.4k$, and $1.4k$ sentences (containing $2.2k$, $0.9k$, and $0.9k$ CSW sentences). 

For Egyptian Arabic-to-English parallel sentences, we obtain 309$k$ parallel sentences from the following parallel corpora: Callhome Egyptian Arabic-English Speech Translation
Corpus \cite{GKA+97,LDC2002T38,LDC2002T39,KCC+14}, LDC2012T09 \cite{ZMD+12}, 
LDC2017T07 \cite{LDC2017T07}, 
LDC2019T01 \cite{LDC2019T01}, 
LDC2021T15 \cite{LDC2021T15}, and MADAR \cite{BHS18}. The corpora cover web (LDC2012T09/LDC2019T01), chat (LDC2017T07/LDC2021T15), and conversational (Callhome/MADAR) domains. We use the corpora data splits if pre-defined, otherwise, we follow the guidelines provided by \newcite{DHR+13}. Data preprocessing is discussed in Appendix~\ref{sec:appendix_data_preprocessing}.

\subsection{Setup of Augmentation Approaches}
Through augmentation, we convert the source side of the 309$k$ Arabic-to-English parallel sentences to CSW Arabic-English. For word alignments, we use Giza++ \cite{och03:asc}.\footnote{
Following \newcite{hamed2022investigating}, in lexical replacements, we take the union of grow-diag-final alignments trained on word and stem spaces. We use the same alignments in linguistic theories, as it produces more generations compared to the default alignment setup used in the GCM tool (grow-diag-final-and alignments trained on word space using fast-align \citep{dyer2013simple}).} 
For the augmentation approaches that require CSW parallel sentences, we utilize ArzEn-ST. 
In \BT, we train the model on the train sets of the parallel corpora outlined in Section \ref{sec:data}, with reversed source and target sides. The predictive model in \LexPred~is trained on the portion of ArzEn-ST train set having CSW sentences. That subset is also utilized in the linguistic theories to obtain the reference SPF distribution ($=0.22$). It is also utilized in \lexDict~and \lexRand, where the value of $x$ is set to 19\% of the source words based on the percentage of English words in ArzEn-ST train set CSW sentences, which is 18.8\%. However, the average percentage calculated over sentences is 22.1\% with a standard deviation of 17.5\%. The decision of 19\% is in agreement with \newcite{hussein2023textual} where the authors report LM perplexities achieved by embedding different percentages of English words in Arabic text using random lexical replacement and decide on a percentage of 20\%. In future work, we believe an interesting direction is to model CSW distribution to obtain a wider coverage of various CSW levels rather than targeting a single percentage for all sentences.

\subsection{Machine Translation System}
We train a Transformer model using Fairseq \cite{OEB+19} on a single GeForce RTX 3090 GPU. We use the hyperparameters from the FLORES benchmark for low-resource machine translation \cite{GCO+19}.\footnote{FLORES hyperparameters outperformed \citet{vaswani2017attention} in \citet{gaser2022exploring} on the same utilized datasets.} The hyperparameters are given in Appendix \ref{sec:appendix-MThyperparameters}. We use a BPE model trained jointly on source and target sides with a vocabulary size of $16k$ (which outperforms $1,3,5,8,32,64k$). The BPE model is trained using Fairseq with character\_coverage set to $1.0$. For MT training data, we use the train sets of the corpora outlined in Section~\ref{sec:data}. For the augmentation experiments, we append the synthetically generated CSW Arabic-English to English parallel sentences. For development and evaluation of the MT models, we use ArzEn-ST dev and test sets.

\section{Evaluation}

In this section, we present intrinsic evaluation, human evaluation, and extrinsic evaluation.
\subsection{Intrinsic Evaluation}
In Table \ref{table:aug_stats}, we report the number of CSW sentences generated per technique as well as CSW statistics. We report that the number of augmentations varies considerably across techniques: 
\lexDict~> \lexRand~> \BT~> {\sc EC}~> \LexPred~> {\sc ML}.

With regards to CSW metrics, we report Code-mixing Index (CMI) \cite{GD16}, SPF, and the average percentage of English tokens over sentences. CMI reflects the level of mixing between multiple languages, and is calculated on the sentence-level as follows:
\begin{align*}
\resizebox{\columnwidth}{!}{$CMI(x)=\frac{\frac{1}{2}*(N(x)-max_{L_i\in \textbf{L}}\{t_{L_i}\}(x))+\frac{1}{2}P(x)}{N(x)}$}
\end{align*}
where $N$ is the number of language-dependent tokens in sentence $x$; $L_i \in \textbf{L}$ is the set of languages in the corpus; $max_{L_i\in \textbf{L}}\{t_{L_i}\}$ is the number of tokens in the dominating language in $x$; 
and $P$ is the number of switch points in $x$, where $0 \leq P < N$. The corpus-level CMI is calculated as the average of sentence-level CMI values.

We observe that in general, \lexRand~and \LexPred~provide the closest figures to ArzEn-ST with regards to CSW metrics. It is to be noted that unlike \lexRand~and SPF-based linguistic theories, no explicit CSW heuristics were provided to \LexPred, and the predictive model learnt to imitate the CSW frequency in ArzEn-ST. In the case of linguistic theories, we note that SPF sampling provides CMI and SPF figures that are closer to ArzEn-ST than random sampling. 
Finally, we report that the linguistic theories and \BT~augmentations contain high percentages of English words. 

\begin{table}[]
\centering
\setlength{\tabcolsep}{3pt}
\begin{tabular}{|l|r|c|cc|c|}
\cline{2-6}
\multicolumn{1}{c|}{}& \multicolumn{1}{c|}{\textbf{Size (\textit{k})}} & \multicolumn{1}{c|}{\textbf{CMI}} & \multicolumn{1}{c}{\textbf{SPF}} & \multicolumn{1}{c|}{\textbf{SPF$_{\sigma}$}} & \multicolumn{1}{c|}{\textbf{\%En}} \\\hline
ArzEn-ST & \multicolumn{1}{c|}{-}& 0.21& 0.22& 0.13& 22.1\\\hline
\lexDict & 239.6& 0.28& 0.33& 0.12& 22.5\\
\lexRand& 192.7& 0.25& 0.24& 0.12& 31.9\\
\LexPred& 112.9& 0.24& 0.22& 0.13& 36.8\\
\ecRand& 142.1& 0.30& 0.29& 0.14& 59.0\\
\ecSPF& 142.1& 0.25& 0.24& 0.08& 64.4\\
\mlRand& 98.2& 0.27& 0.27& 0.14& 60.8\\
\mlSPF& 98.2& 0.25& 0.25& 0.10& 63.1\\
\BT& 151.1& 0.18& 0.19& 0.14& 65.2\\\hline
\end{tabular}
\caption{
The number of generated sentences per technique, and their CMI and SPF mean and standard deviation (SPF/SPF$_{\sigma}$) values and average percentage of English words (\%En). We also report the figures for the CSW sentences in ArzEn-ST train set as reference.}
\label{table:aug_stats}
\end{table}

\subsection{Human Evaluation}
\label{sec:human_eval}
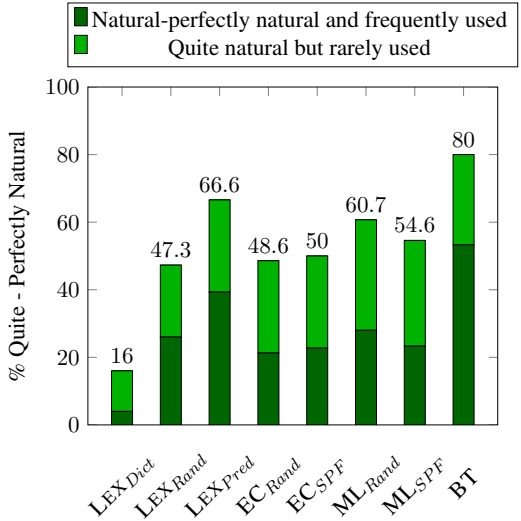
\begin{figure}
\centering
\resizebox{0.9\columnwidth}{!}{%
\begin{tikzpicture}
\begin{axis}[
    ybar stacked,
    ymin=0,
    ymax=100,
    xtick=data,
    legend style={at={(0.5,1.25)},anchor=north,legend columns=1},
    ylabel={\% Quite - Perfectly Natural},
    reverse legend=false, 
    xticklabels from table={\testdata}{Label},
    xticklabel style={text width=2cm,align=center,rotate=45}
]
\addlegendentry{Natural-perfectly natural and frequently used}
\addplot [fill=black!60!green] table [y=series1, meta=Label, x expr=\coordindex] {\testdata};
\addlegendentry{Quite natural but rarely used}
\addplot [fill=black!30!green] table [y=series2, meta=Label, x expr=\coordindex] {\testdata};

\addplot [
    ybar, 
    nodes near coords,
    nodes near coords style={%
        anchor=south,%
    },
] table [ y expr=0.00001, x expr=\coordindex] {\testdata};

\end{axis}
\end{tikzpicture}
}
\caption{The percentage of augmentations with 3$\leq$~MOS~$\leq$5 (quite natural but rarely used - perfectly natural and frequently used) per technique.}
\label{fig:human_eval}
\end{figure}

In order to assess the quality of the synthetically generated CSW sentences, we perform a human evaluation study. Out of the original sentences that get augmented by all techniques, we randomly sample 150 sentences.\footnote{The sentences are sampled uniformly across the six corpora used in data augmentation to have equal representation of the different domains (web/chat/conversational).} These sentences are evaluated by three annotators across the eight augmentation techniques against two measures: understandability and naturalness. All three annotators are female Egyptian Arabic-English bilingual speakers, in the age range of 33-39, all graduates of private English schools. We follow the rubrics introduced by \newcite{PC21}, outlined in Table~\ref{table:human_eval_rubrics}. Understandability is rated on a scale of 1-3 and naturalness is rated on a scale of 1-5 where scores of 3-5 are assigned to natural sentences with different levels of commonality to be encountered in real life. A total of 1,200 augmentations are annotated by each of the three annotators for both understandability and naturalness, giving a total of 7,200 annotations.\footnote{The annotation task took an average of 9 hours per annotator, and each annotator was paid \$160.} For each augmentation, we calculate the mean opinion score (MOS) as the average of scores received by the three annotators. The full results are provided in Appendix \ref{sec:appendix-humanEval}, where the percentage of sentences falling under each MOS range per technique is presented in Table \ref{table:MOS_scores}. In Figure \ref{fig:human_eval}, we show the percentage of sentences perceived as natural by annotators across techniques (summation of the last two rows in Table \ref{table:MOS_scores}). 
We observe the following ranking between techniques: 
\BT~> \LexPred~> {\sc ML} > {\sc EC} > \lexRand~> \lexDict.

\begin{table}[t]
\centering
\setlength{\tabcolsep}{3pt}
\begin{tabular}{ | c | p{7cm} | }
\hline
\multicolumn{2}{|c|}{\textbf{Understandability}}\\\hline
1  & No, this sentence doesn't make sense. \\
2  & Not sure, but I can guess the meaning of this sentence. \\
3 &  Certainly, I get the meaning of this sentence. \\\hline 
\multicolumn{2}{|c|}{\textbf{Naturalness}}\\\hline
1 & Unnatural, and I can't imagine people using this style of code-mixed Arabic-English. \\
2 & Weird, but who knows, it could be some style of code-mixed Arabic-English.\\
3 & Quite natural, but I think this style of code-mixed Arabic-English is rare.\\ 
4 & Natural, and I think this style of code-mixed Arabic-English is used in real life.\\
5 & Perfectly natural, and I think this style of code-mixed Arabic-English is very frequently used.\\\hline
\end{tabular}
\caption{The evaluation dimensions for human evaluation, following \newcite{PC21}.}
\label{table:human_eval_rubrics}
\end{table}

With regards to linguistic theories, as noted by \newcite{dougruoz2023survey}, computational implementations of linguistic theories do not necessarily generate natural CSW sentences that would mimic human CSW generation. We elaborate on this point in Section \ref{sec:aug_insights}. While {\sc ML} achieves higher naturalness ratings than {\sc EC}, we do not observe superiority across the different sampling techniques, which can be due to the SPF values only changing slightly between both techniques in our case. This can be different in other setups with different reference SPF distributions. With regards to understandability, there is less variability across the techniques (91-96\% of the augmentations are given ratings between 2 and 3), except for \lexDict~(the percentage is 65\%). We perform inter-annotator agreement by applying pairwise Cohen Kappa \cite{Cohen:1960:coefficient}, reporting 0.25-0.28 (fair agreement) on naturalness between annotator pairs. Low agreement on this task is expected, as CSW attitude is speaker-dependent \cite{vu2013investigation}. The pairwise Cohen Kappa scores for understandability are higher (0.33-0.35), yet still showing fair agreement. We also apply Fleiss' Kappa \cite{fleiss1971measuring} across all annotators, scoring fair agreement of 0.312 and 0.249 for   understandability and naturalness.\footnote{We use the implementation provided in: \\\url{https://github.com/Shamya/FleissKappa/blob/master}}

 \pgfplotstableread{
Label   series1 series2
{\lexDict}  4.0     12.0  
{\lexRand}  26.0    21.3
{\LexPred}  39.3    27.3
{\ecRand}   21.3    27.3
{\ecSPF}    22.7    27.3
{\mlRand}   28.0    32.7
{\mlSPF}    23.3    31.3
{\BT}       53.3    26.7
    }\testdata

\subsection{Extrinsic MT Evaluation}
The augmentation techniques covered in this study vary in terms of requirements. One main difference is the reliance on CSW parallel data, which is only available for a few CSW language pairs \cite{hamed2022arzenST}. To have a fair comparison and to show the effectiveness of the techniques in both cases (availability and lack of CSW-focused parallel corpora), we run two sets of experiments:
\begin{itemize}
    \item Zero-shot setting: In this setting, our baseline system is trained only using the 309$k$ monolingual Arabic-to-English parallel sentences. We extend the training data with augmentations generated using  techniques that do not require CSW parallel data, namely: \lexDict, \lexRand, {\sc EC}, and {\sc ML}.
    \item Non-zero-shot setting: In this setting, we assume the availability of CSW parallel data. We train our baseline system using the monolingual Arabic-to-English parallel sentences in addition to ArzEn-ST corpus. We then append the augmentations generated by each of the investigated techniques.
\end{itemize}

In the following sections, we present our baseline systems and the results for zero-shot and non-zero-shot settings. The full results are reported in Table \ref{table:extrinsic_eval_results}, showing BLEU \cite{PRT+02}, chrF, chrF++ \cite{popovic2017chrf++}, and BERTScore (F1) \cite{ZKW+19}. BLEU, chrF and chrF++ are calculated using SacreBLEU \cite{Pos18}. We report performance on ArzEn-ST test set; on all sentences as well as CSW sentences only. Our analysis in this section is based on chrF++. This choice is based on chrF++ showing higher correlation with human judgments over chrF \cite{popovic2017chrf++} and chrF showing higher correlation over BLEU \citep{Kocmi2021ship}.  We report performance on ArzEn-ST test set CSW sentences, as this is our main concern. Statistical significance tests for zero- and non-zero-shot settings are shown in Table \ref{table:stat_sig}.

\subsubsection{Baselines}
\label{sec:baseline}
We develop the following MT baselines, showing the improvements achieved by each source of data:
\begin{itemize}
    \item BL$_{CSW}$: We train it solely on ArzEn-ST train set, having 3.3$k$ parallel sentences.
    \item BL$_{Mono}$: We train it on the 309$k$ monolingual Arabic-to-English parallel sentences.
    \item BL$_{MonoTgt}$: In BL$_{Mono}$, we observe that English words on the source side get dropped in translation. This issue has been previously tackled by researchers using techniques including direct copying \cite{SZY+19} or the use of a pointer network \cite{MLJ+19}. We propose a simple technique of including target-target pairs in the training process. In other words, in addition to the source-target sentences used in BL$_{Mono}$, we append the English (target) sentences on both source and target sides, ending up with 617$k$ parallel sentences. Our hypothesis is that by doing so, the model learns to retain the English words on the source side through translation.
    \item BL$_{All}$: We include the same data as in BL$_{MonoTgt}$, in addition to ArzEn-ST train set, giving a total of 620$k$ parallel sentences.
\end{itemize}

\begin{figure}[t]
\centering
\resizebox{0.85\columnwidth}{!}{
\begin{tikzpicture}
\begin{axis}[ybar, symbolic x coords={BL$_{CSW}$, BL$_{Mono}$, BL$_{MonoTgt}$, BL$_{All}$}, enlarge x limits=0.2, ylabel={chrF++}] 
\addplot+ coordinates {(BL$_{CSW}$, 27.1) (BL$_{Mono}$, 39.4) (BL$_{MonoTgt}$, 55) (BL$_{All}$, 57.3)};
\end{axis}
\node (n1) at (1,0.7) {27.1};
\node (n2) at (2.6,2.7) {39.4};
\node (n3) at (4.2,5.1) {55.0};
\node (n4) at (5.9,5.5) {57.3};
\draw (n1) (n2) (n3) (n4);
\end{tikzpicture}
}
\caption{chrF++ scores of the different baselines on ArzEn-ST test set CSW sentences. }
\label{fig:baselines}
\end{figure}
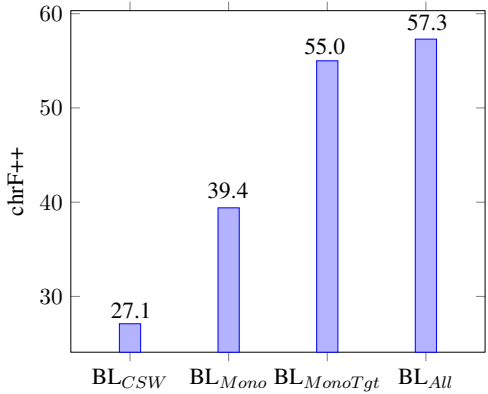

The chrF++ scores are shown in Figure \ref{fig:baselines} (full results in Table \ref{table:extrinsic_eval_results} Exp~1-4). The effectiveness of 
the simple step of 
adding target-target pairs during training is confirmed, where BL$_{MonoTgt}$ achieves an increase of +15.6 chrF++ points over BL$_{Mono}$. Adding ArzEn-ST train set (BL$_{All}$) results in further +2.3 chrF++ points, achieving 57.3 on chrF++.

\subsubsection{Zero-shot Setting Experiments}
\label{sec:zero-shot_setting}
This setting is tailored to the majority of CSW language pairs, that are under-resourced and lack CSW-focused parallel corpora. We demonstrate the effectiveness of the augmentation techniques in a zero-shot setting. Given that \LexPred~and \BT~are reliant on CSW parallel data, they are excluded from this comparison. We include the following approaches: \lexDict, \lexRand, \ecRand, \ecSPF, \mlRand,~and \mlSPF. We acknowledge that some of these approaches rely on heuristics obtained from CSW data, such as SPF value or the enforced CSW percentage. However, we argue that these figures can be obtained from textual data (that is more easily accessible than parallel data). 
The baseline in this setting is BL$_{MonoTgt}$, which is our best baseline that does not utilize real CSW data.

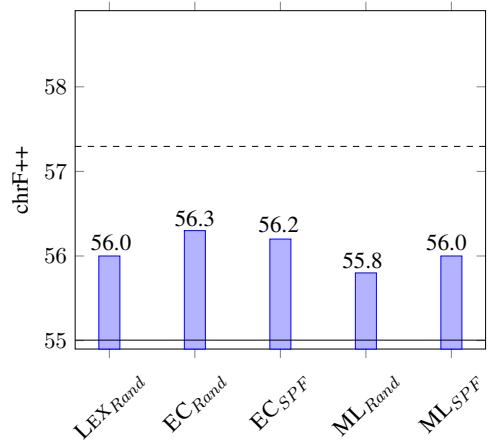
\begin{figure}[t]
\centering
\resizebox{0.85\columnwidth}{!}{%
\begin{tikzpicture}
\begin{axis}[ybar, symbolic x coords={\lexRand, \ecRand, \ecSPF, \mlRand, \mlSPF}, ylabel={chrF++},
ymin=54.9,
ymax=58.9,
xticklabel style={rotate=45}
]
\addplot+ coordinates {(\lexRand, 56) (\ecRand, 56.3) (\ecSPF, 56.2) (\mlRand, 55.8) (\mlSPF, 56)};
\end{axis}
\node (n2) at (0.6,1.8) {56.0};
\node (n3) at (2,2.2) {56.3};
\node (n4) at (3.4,2.1) {56.2};
\node (n5) at (4.8,1.5) {55.8};
\node (n6) at (6.2,1.8) {56.0};
\draw (n2) (n3) (n4) (n5) (n6);
\draw (0,0.15) -- (6.85,0.15);
\draw [dashed] (0,3.41) -- (6.85,3.41);
\end{tikzpicture}
}
\caption{The effectiveness of the augmentation techniques in a zero-shot setting. We show the chrF++ scores on ArzEn-ST test set CSW sentences. The solid and dashed lines represent BL$_{MonoTgt}$ and BL$_{All}$.}
\label{fig:zero-shot}
\end{figure}

We report that \lexDict~degrades the MT performance, falling 3.2 chrF++ points below the baseline. We present the chrF++ scores for the other techniques in Figure \ref{fig:zero-shot} (full results in Table~\ref{table:extrinsic_eval_results} Exp~5-10). We observe that linguistic-based models and \lexRand~perform equally well, despite \lexRand~ generating more data. As shown in Table \ref{table:stat_sig}, there is no statistical significance between \lexRand~and linguistic-based models. 
Comparing the linguistic theories, EC performs better than ML, however, there is no difference between SPF and random sampling strategies. Overall, \ecRand~performs the best, with statistical significance over \mlRand~and \mlSPF,~achieving +1.3 chrF++ points over BL$_{MonoTgt}$.

\subsubsection{Non-zero-shot Experiments}
\label{sec:non-zero-shot_setting}
\begin{figure}[t]
\centering
\resizebox{0.85\columnwidth}{!}{%
\begin{tikzpicture}
\begin{axis}[
    ybar, symbolic x coords={\lexRand, \LexPred, \ecRand, \ecSPF, \mlRand, \mlSPF, \BT}, 
    ylabel={chrF++},
    ymin=54.9,
    ymax=58.9,
    xticklabel style={rotate=45}
]
\addplot+ coordinates {(\lexRand, 57.5) (\LexPred, 58) (\ecRand, 56.9) (\ecSPF, 57.3) (\mlRand, 57.6) (\mlSPF, 57.5) (\BT, 58.6)};
\end{axis}
\node (n2) at (0.6,3.9) {57.5};
\node (n3) at (1.5,4.6) {56.9};
\node (n4) at (2.5,3.1) {57.3};
\node (n5) at (3.4,3.6) {57.6};
\node (n6) at (4.4,4.1) {57.5};
\node (n7) at (5.3,3.9) {58.0};
\node (n8) at (6.3,5.5) {58.6};
\draw (n2) (n3) (n4) (n5) (n6) (n7) (n8);]
\draw [dashed] (0,3.41) -- (6.85,3.41);
\draw (0,0.15) -- (6.85,0.15);
\end{tikzpicture}
}
\caption{The effectiveness of the augmentation techniques in a non-zero-shot setting. We show the chrF++ scores on ArzEn-ST test set CSW sentences. The solid and dashed lines represent BL$_{MonoTgt}$ and BL$_{All}$.}
\label{fig:non-zero-shot}
\end{figure}
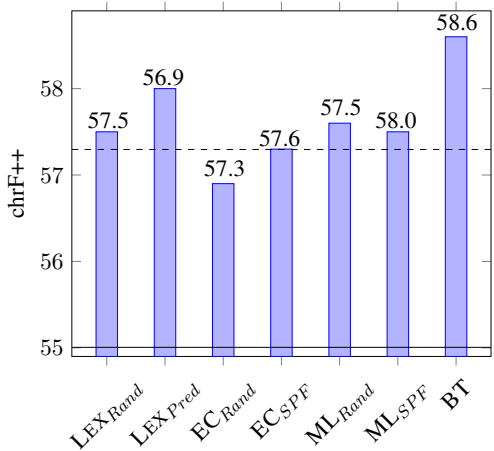

In this setting, we assume the availability of CSW-focused parallel data, and thus compare all augmentation techniques. The baseline for this setting is BL$_{All}$. The chrF++ scores are shown in Figure \ref{fig:non-zero-shot} (full results in Table \ref{table:extrinsic_eval_results} Exp~11-18).\footnote{The lexical replacements results differ from \newcite{hamed2022investigating} due to the following design decisions: (1) we append target-target pairs in the training data; and (2) we only include generated CSW sentences, and not sentences that get fully converted to  English, in order to better control for variables.}

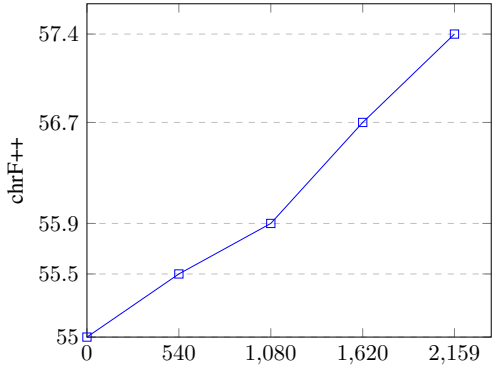
\begin{figure}[t]
\centering
\resizebox{0.85\columnwidth}{!}{%
\begin{tikzpicture}
\begin{axis}[
    ylabel={chrF++},
    xmin=0, ymin=55,
    xtick={0,540,1080,1620,2159},
    ytick={55,55.5,55.9,56.7,57.4},
    ymajorgrids=true,
    grid style=dashed,
]

\addplot[
    color=blue,
    mark=square,
    ]
    coordinates {(0,55) (540,55.5) (1080,55.9) (1620,56.7) (2159,57.4)};
    
\end{axis}
\end{tikzpicture}
}
\caption{Learning curve of adding different amounts of ArzEn-ST train set CSW sentences to the BL$_{MonoTgt}$ training data. We show the
chrF++ scores on ArzEn-ST test set CSW sentences.}
\label{fig:arzen_learningCurve}
\end{figure}

\lexDict~ falls below BL$_{All}$ by 1.4 chrF++ points, we thus exclude it from Figure~\ref{fig:non-zero-shot}. We observe that \LexPred~and \BT~outperform \lexRand~and linguistic theories. The best performance is achieved by \BT, achieving +1.3 chrF++ points over BL$_{All}$. 
We also report that \lexRand~and linguistic theories are unable to achieve significant improvements over BL$_{All}$.\footnote{The improvement achieved by \mlRand~over BL$_{All}$ is not statistically significant.} We examine the amount of real in-domain CSW data that would result in equivalent performance achieved by \lexRand~and linguistic theories in the zero-shot setting. In Figure \ref{fig:arzen_learningCurve}, we show a learning curve by adding different amounts of ArzEn-ST train set CSW sentences to BL$_{MonoTgt}$ training data, and show that \lexRand~and linguistic theories (generating 98-192$k$ CSW synthetic sentences) perform on par at 50\% of ArzEn-ST train set CSW sentences ($\approx$ 1,080 sentences).

\section{Discussion}
\label{sec:discussion}
In this section, we revisit our RQs:
\paragraph{RQ1 - Which augmentation techniques perform the best for MT?} 
In the zero-shot setting, \lexRand~and linguistic theories achieve similar performance, with {\sc EC} outperforming {\sc ML} models. In the non-zero shot setting, \BT~outperforms all techniques, followed by \LexPred. Both techniques, being trained on real CSW data, are able to generate more natural CSW sentences, that could also be closer in CSW style to ArzEn-ST.
\paragraph{RQ2 - Does generating more natural synthetic CSW sentences entail improvements in MT?} Here, we look into the relation between MT scores and naturalness ratings. 
In the non-zero shot setting, we report a correlation of 0.97 between the chrF++ scores (presented in Figure \ref{fig:non-zero-shot}) and the percentage of sentences perceived as natural (presented in Figure \ref{fig:human_eval}). This demonstrates a strong positive correlation between MT performance and naturalness of augmentations.

Given that the number of augmentations varies considerably across techniques (shown in Table~\ref{table:aug_stats}), this variation could empower some techniques over others, affecting performance as an effect of quantity rather than quality. Therefore, we perform another set of experiments where we control for this variable. We report results under a constrained setup, where we restrict the augmentations appended to the baseline training data to only those that are successfully augmented across all techniques (= 24.8$k$ sentences). We first append the constrained augmentations per technique to BL$_{MonoTgt}$ training data. The results are presented in Table \ref{table:extrinsic_eval_results} Exp~19-26. The order based on chrF++ is:
\BT~> \LexPred~> [\lexRand~\& linguistic theories] > \lexDict. The correlation between the chrF++ scores achieved on ArzEn-ST test set CSW sentences and the percentage of sentences perceived as natural is 0.95. 
We replicate the constrained experiments with appending the constrained augmentations to BL$_{All}$ training data. Given the availability of CSW data in the training data, and with constraining the amounts of augmented data, the majority of the models show no improvements over BL$_{All}$. We therefore cannot use this setup to make conclusions on the relation between quality and performance. However, from the previously discussed findings, we confirm a positive relation between the naturalness of generated synthetic sentences and MT performance.

\section{Insights into Augmentations}
\label{sec:aug_insights}
In this section, we present insights into the augmentations produced by the different techniques, further elaborating on their strengths and weaknesses. All examples mentioned in this section refer to the examples demonstrated in Table~\ref{table:aug_examples}.

\paragraph{Lexical Replacements:} The main drawback in \lexDict~is that the replaced words might not be correct translations within context, which can negatively affect the MT model. As shown in Table~\ref{table:aug_examples} Example~1, \<اولع> \textit{AwlE}\footnote{Buckwalter Arabic Transliteration \cite{Habash:2007:arabic-transliteration}.} `turn on', in the context of \textit{turn on this light}, is replaced by `kindle'. This drawback is also observed in the case of ambiguous words, as shown in Example 2, where the word \<طابع> \textit{TAbE} `stamp' is replaced by `impression'. With regards to \lexRand, CSW can occur at unnatural locations, such as replacing \<ده> \textit{dh} `this' in Example~1. This is less likely for \LexPred, which is reflected in human evaluation.

\paragraph{Linguistic theories:} 
We observe that applying linguistic theories does not guarantee naturalness, e.g., the augmentation provided by \ecSPF~shown in Example 3, despite being a correct augmentation under the EC theory, was given a rating of `2' by all annotators. Moreover, the effectiveness of these techniques is tied to (and currently restricted by) the performance of the available tools that implement them. We observe that the augmentations obtained from the GCM tool in some cases violate the EC or ML theories. For the EC theory, in Example 4, we demonstrate a case where Arabic-to-English alternation occurs at the word `station' which is a point of syntactic divergence in \<محطة الاوتوبيس> \textit{mHTp AlAwtwbys} and `the coach station'. For the ML theory, in Example 5, the augmentation includes the stand-alone CSW segment `in', while replacements of closed-class constituents (including prepositions) is prohibited. As the tool relies on generating Arabic parse trees from English parse trees using alignments, errors are likely to be introduced. Furthermore, as noted by \newcite{hussein2023textual}, the augmentations are sometimes missing information from the original sentences.\footnote{We reduce the effect of this issue by validating that the generated sentences are complete using the alternational matrices computed by the tool in the generation process, and giving priority to sampling from validated augmentations.}

\paragraph{BT:} We observe that \BT~is capable of generating correct morphological code-switching (MCS). As shown in Example 6, the MCS construction `\<بت>+handle' \textit{bt+handle} `handles' is correctly composed of 
\<بت> \textit{bt} `progressive-imperfect-2nd-masc-sing' preceding the verb `handle' and `field' is correctly preceded by the definite article \<ال> \textit{Al} `the'. While researchers have provided insights into common Arabic-English MCS constructs 
\cite{kniaz2017english,kniaz2021embedded,HDL+22}, there is no current research that allows for modeling Arabic-English MCS in a rule-based approach. Therefore, the ability of neural-based approaches to generate MCS is an advantage. On the other hand, similar to the partial transcription issue noted in \newcite{chowdhury2021towards} for ASR models using BPE, the \BT~approach can provide partial translations of words, such as `modifications' translated to s+\<تعديل> \textit{tEdyl+s} `modification+s'(Example~7). \BT~ might also provide literal translation. With both issues combined, we find cases such as `locker' being translated to er+\<قفل> \textit{qfl+er} `lock+er'.

\section{Conclusion and Future Work}
We present a comparative study between different CSW data augmentation techniques and their effectiveness for MT in both zero-shot and non-zero-shot settings. We show that in the zero-shot setting, random lexical replacement performs equally well as linguistic theories. In the case of non-zero shot setting, back-translation performs best, followed by CSW predictive-based lexical replacement. Both approaches also stand out in human evaluation, where we confirm a positive correlation between naturalness of augmentations and MT performance. However, both approaches are reliant on expensive and limited CSW parallel data. Overall, the set of approaches examined proves useful in alleviating data scarcity. Each approach comes with particular merits and requirements, guiding the choice for different research needs. In future work, we plan on enhancing the back-translation approach to leverage larger amounts of English data. In parallel, we will investigate the effectiveness of generative AI to broaden the benchmark of approaches, and expand our study to cover other NLP tasks.

\section*{Limitations}
One limitation of the presented work is that the models were evaluated on one test set only, and therefore, we cannot interpret how the models will perform on other sets, covering other domains and sources (spoken versus written). Another limitation is that the study involves only one language pair. Further research is needed to investigate whether the findings hold for other language pairs. A third limitation is the low variability in the annotators' demographics, as the three annotators are female annotators, in the same age group, receiving similar levels of education. Including a broader set of annotators would enrich research with insights on the level of agreement between annotators with wider background differences.
\section*{Ethics Statement}
We could not identify any ethical issues in the work, and to our best knowledge, we believe it complies with the ACL Ethics Policy. We use ArzEn-ST corpus, which is distributed under an Attribution-ShareAlike 4.0 International license, where we adhere to its intended usage. All other parallel corpora are also publicly available, including MADAR, Callhome, and LDC corpora. 

\bibliography{main}

\appendix

\section{Augmentation Examples}
\label{appendix:aug_examples}
In Table~\ref{table:aug_examples}, we present examples of augmentations generated by the different techniques. These examples are discussed in Section \ref{sec:aug_insights}. 
\begin{table*}[th!]
    \centering
    \begin{tabular}{|l|l|r|}
    \hline
    \multicolumn{3}{|c|}{\textbf{Examples}}\\\hline
        &Src        & \<ازاي اولع النور ده ?>\\
        &Tgt        & how can i turn on this light ?\\
        &\lexDict    & \<النور ده ?> kindle \<ازاي>\\
        &\lexRand    & \<?> this \<ازاي اولع النور> \\
        1&\LexPred    & \<ده ?> light \<ازاي اولع>\\
        &\ecRand     &  \<النور ده ?> how can i turn on\\
        &\ecSPF      &  \<?> this light \<ازاي اولع >\\
        &\mlRand     &  \<ده ?> light \<ازاي اولع>\\
        &\mlSPF      &  \<النور ?> how can i turn on this\\
        &\BT         &  \<ده ?> light   \<ازاي اقدر افتح ال>\\
        \hline
        &Src         & \< تاني ?> \<طابع> \<جبت> \\
        2&Tgt         & got another stamp ?\\
        &\lexDict    &  \< تاني ?> impression \<جبت> \\
        \hline
        &Src        & \<لازم ناخد جواب من الادارة في طنطا كل شهرين>\\
        3&Tgt        & we must take a letter from the management in tanta ; every two months\\
        &\ecRand         & \<شهرين> in tanta ; every \<لازم ناخد جواب من الادارة>\\
        \hline
        &Src        & \<هو محطة الاوتوبيس فين ?>\\
        4&Tgt        & where 's the coach station ?\\
        &\ecRand         & \<الاوتوبيس فين ?> station \<هو>\\
        \hline
        &Src        & \<انا سبت محفظتي في محلك .>\\
        5&Tgt       & i left my wallet in your shop .\\
        &\mlRand    &  \<محلك .> in \<محفظتي> left \<انا>\\    
        \hline
        &Src        & \<لو انت مبتتعاملش مع المجال ده , ممكن تقترح محل متخصص في ده ?>\\
        6&Tgt        & if you don 't handle that field , could you suggest a store specializing in it ?\\
        &\BT        & \<ده , ممكن تقترح محل متخصص فيه ?> field \<ال> handle \<لو انت مش بت>\\
        \hline
        &Src        & \<فيه اي تعديلات انت عايز تعملها ?>\\
        7&Tgt        & are there any modifications you would like to make ?\\
        &\BT        & \<تحب تعملها ?> s\<في اي تعديل>\\
        \hline
    \end{tabular}
    \caption{Examples of synthetic CSW sentences generated by the different augmentation techniques, demonstrating strengths and weaknesses of techniques. Given that Arabic is written from right to left, we display all augmentations in a right-to-left orientation.}
    \label{table:aug_examples}
\end{table*}

\section{Data Preprocessing}
\label{sec:appendix_data_preprocessing}
Following \newcite{hamed2022investigating}, we remove corpus-specific annotations, remove URLs and emoticons through tweet-preprocessor, tokenize numbers, apply lowercasing, run Moses’ \cite{KHB+07} tokenizer as well as MADAMIRA \cite{PAD+14} simple tokenization (D0), and perform Alef/Ya normalization.\footnote{\url{https://pypi.org/project/tweet-preprocessor/}} For entries with words having literal and intended translations, we opt for one translation having all literal translations and another having all intended translations. For LDC2017T07, we utilize the work by \newcite{SUH20}, where the authors used a sequence-to-sequence model to transliterate the corpus text from Arabizi (where Arabic words are written in Roman script) to Arabic orthography. For the Egyptian Arabic-to-English parallel corpora discussed in Section \ref{sec:data}, we only utilize the 309k monolingual Egyptian Arabic-to-English parallel sentences available in these corpora, where we do not utilize the parallel sentences with code-switching within the scope of this work. In future work, it would be interesting to investigate how the effectiveness of data augmentation varies with the availability of different amounts of real CSW parallel data, to draw further conclusions under different levels of low-resourcefulness. Also, for MADAR and LDC2012T09, we only utilize the Egyptian Arabic subsets of both corpora.

\section{Hyperparameters}
\label{sec:appendix-MThyperparameters}
For finetuning mBERT for the CSW predictive model, we set the epochs to 5, drop-out rate to 0.1, warmup steps to 500, batch size to 13, and learning rate to 0.0001. The training and inference time took $\approx$ 12 hours.

For MT, we use the following train command: python3 fairseq\_cli/train.py \${DATA\_DIR} --source-lang src --target-lang tgt --arch transformer --share-all-embeddings --encoder-layers 5 --decoder-layers 5 --encoder-embed-dim 512 --decoder-embed-dim 512 --encoder-ffn-embed-dim 2048 --decoder-ffn-embed-dim 2048 --encoder-attention-heads 2 --decoder-attention-heads 2 --encoder-normalize-before --decoder-normalize-before  --dropout 0.4 --attention-dropout 0.2 --relu-dropout 0.2  --weight-decay 0.0001 --label-smoothing 0.2 --criterion label\_smoothed\_cross\_entropy --optimizer adam --adam-betas '(0.9, 0.98)' --clip-norm 0 --lr-scheduler inverse\_sqrt --warmup-updates 4000 --warmup-init-lr 1e-7  --lr 1e-3 --stop-min-lr 1e-9 --max-tokens 4000 --update-freq 4 --max-epoch 100 --save-interval 10 --ddp-backend=no\_c10d
\section{MT Results}
\label{sec:appendix_mt_results}
In Table \ref{table:extrinsic_eval_results}, we report the MT results, showing BLEU, chrF, chrF++, and BERTScore(F1). The statistical significance between the models in the zero-shot and non-zero-shot settings for chrF++ achieved on ArzEn-ST test set CSW sentences are shown in Table \ref{table:stat_sig}. The number of parameters in the models for Exp~1 is 39,712,768 and Exp~2-26 is 44,967,936. The training time taken by Exp~1 is $\approx$ 8 minutes, Exp~2 $\approx$ 2.6 hours, and Exp~3-26 $\approx$ 5.2-6.5 hours. 

\begin{table*}[th!]
\centering
\setlength{\tabcolsep}{2pt}
\begin{tabular}{| l | l | r | c  c  c  c  | c  c  c  c |}
\cline{4-11}
\multicolumn{3}{c|}{} &\multicolumn{4}{c|}{\textbf{All Test Sentences}} &\multicolumn{4}{c|}{\textbf{\textbf{CSW Test Sentences}}} \\\hline
\multicolumn{1}{|c|}{\textbf{Exp}}&\multicolumn{1}{c|}{\textbf{Model}}& \multicolumn{1}{c|}{$|$\textbf{Train}$|$} & \textbf{BLEU} &\textbf{chrF} & \textbf{chrF++} &\textbf{BertScore(F1)} & \textbf{BLEU} &\textbf{chrF} & \textbf{chrF++} &\textbf{BertScore(F1)}\\
\hline
\hline
\multicolumn{11}{|c|}{\textbf{Baselines}}\\\hline
1&BL$_{CSW}$ &3,340&8.3&27.2&26.6&0.218&8.3&27.8&27.1&0.175\\
2&BL$_{Mono}$ &308,689&22.2&42.1&41.4&0.387&20.7&39.9&39.4&0.315\\
3&BL$_{MonoTgt}$ &617,378&31.7&54.9&53.5&0.519&32.8&56.5&55.0&0.510\\
4&BL$_{All}$ &620,718&\textbf{34.4}&\textbf{57.4}&\textbf{55.7}&\textbf{0.547}&\textbf{35.6}&\textbf{59.1}&\textbf{57.3}&\textbf{0.549}\\
\hline\hline
\multicolumn{11}{|c|}{\textbf{Zero-shot Experiments}}\\\hline
5&+{\lexDict} &857,022&29.8&52.3&51.1&0.499&30.2&53.1&51.8&0.474\\
6&+{\lexRand} &810,030&32.8&55.7&54.3&\textbf{0.531}&34.1&57.5&56.0&\textbf{0.529}\\
7&+\ecRand &759,478&\textbf{33.7}&\textbf{56.1}&\textbf{54.7}&0.528&\textbf{34.9}&\textbf{57.8}&\textbf{56.3}&0.522\\
8&+\ecSPF &759,478&33.1&55.8&\textbf{54.5}&0.530&34.5&\textbf{57.6}&\textbf{56.2}&0.527\\
9&+\mlRand &715,610&32.6&55.5&54.2&0.527&33.9&57.2&55.8&0.520\\
10&+\mlSPF &715,610&33.0&55.8&54.4&0.529&34.2&57.4&56.0&0.523\\
\hline\hline
\multicolumn{11}{|c|}{\textbf{Non-zero-shot Experiments}}\\\hline
11&+{\lexDict} &860,362&33.6&56.0&54.5&0.536&34.8&57.6&55.9&0.530\\
12&+{\lexRand} &813,370&34.2&57.1&55.5&0.546&35.9&59.2&57.5&0.546\\
13&+{\LexPred} &733,660&35.2&57.5&56.1&\textbf{0.550}&36.8&59.5&58.0&0.551\\
14&+\ecRand &762,818&33.5&56.6&55.1&0.544&34.9&58.6&56.9&0.547\\
15&+\ecSPF &762,818&34.6&57.0&55.5&0.547&36.2&59.0&57.3&0.549\\
16&+\mlRand &718,950&34.9&57.4&55.8&0.548&36.3&59.3&57.6&0.549\\
17&+\mlSPF &718,950&34.3&57.3&55.7&0.548&35.7&59.2&57.5&0.547\\
18&+\BT &771,793&\underline{\textbf{35.8}}&\underline{\textbf{58.2}}&\underline{\textbf{56.6}}&\underline{\textbf{0.550}}&\underline{\textbf{37.5}}&\underline{\textbf{60.3}}&\underline{\textbf{58.6}}&\underline{\textbf{0.553}}\\
\hline\hline
\multicolumn{11}{|c|}{\textbf{Constrained Experiments}}\\\hline
19&+{\lexDict} &642,221&30.4&53.3&52.0&0.502&31.2&54.5&53.0&0.482\\
20&+{\lexRand} &642,221&32.2&55.3&53.9&0.529&33.3&56.9&55.5&0.524\\
21&+{\LexPred} &642,221&32.9&55.8&54.3&0.530&34.3&57.6&56.1&0.527\\
22&+\ecRand & 642,221&32.2&55.5&54.0&0.525&33.6&57.4&55.7&0.521\\
23&+\ecSPF &642,221&32.7&55.3&53.9&0.526&34.1&57.2&55.6&0.520\\
24&+\mlRand &642,221&32.3&55.3&53.9&0.524&33.4&56.9&55.4&0.517\\
25&+\mlSPF &642,221&32.5&55.3&53.9&0.523&33.9&57.0&55.5&0.518\\
26&+\BT &642,221&\textbf{34.3}&\textbf{56.4}&\textbf{55.0}&\textbf{0.534}&\textbf{36.1}&\textbf{58.4}&\textbf{56.9}&\textbf{0.531}\\
\hline
\end{tabular}
\caption{We report the MT results (BLEU, chrF, chrF++, and BertScore) on ArzEn-ST test set, for all sentences as well as CSW sentences only. We report the results of the baselines (Section \ref{sec:baseline}), zero-shot (Section \ref{sec:zero-shot_setting}), non-zero-shot (Section \ref{sec:non-zero-shot_setting}), and constrained (Section \ref{sec:discussion}) settings. The best performing models in each setting are bolded. The overall best performing model is underlined.} 
\label{table:extrinsic_eval_results}
\end{table*}

\begin{table*}[]
\begin{subtable}{\textwidth}
\centering
\begin{tabular}{|l|c|lllll|}
\cline{3-7}
\multicolumn{2}{c|}{}& \lexDict & \lexRand & \ecRand & \ecSPF & \mlRand \\\cline{2-7}
\multicolumn{1}{c|}{}& chrF++ & \multicolumn{1}{c}{51.8} & \multicolumn{1}{c}{56.0} & \multicolumn{1}{c}{56.3} & \multicolumn{1}{c}{56.2} & \multicolumn{1}{c|}{55.8} \\\hline
\lexDict  & 51.8& \cellcolor[HTML]{EFEFEF} & \cellcolor[HTML]{EFEFEF}            & \cellcolor[HTML]{EFEFEF} & \cellcolor[HTML]{EFEFEF} & \cellcolor[HTML]{EFEFEF} \\
\lexRand  & 56.0 & 0.0010* & \cellcolor[HTML]{EFEFEF} & \cellcolor[HTML]{EFEFEF} & \cellcolor[HTML]{EFEFEF} & \cellcolor[HTML]{EFEFEF} \\
\ecRand & 56.3 & 0.0010* & {\color[HTML]{B42419} 0.0539} & \cellcolor[HTML]{EFEFEF} & \cellcolor[HTML]{EFEFEF} & \cellcolor[HTML]{EFEFEF} \\
\ecSPF & 56.2 & 0.0010* & {\color[HTML]{B42419} 0.1139} & {\color[HTML]{B42419} 0.2208} & \cellcolor[HTML]{EFEFEF} & \cellcolor[HTML]{EFEFEF}            \\
\mlRand & 55.8 & 0.0010* & {\color[HTML]{B42419} 0.0959} & 0.0030* & 0.0120* & \cellcolor[HTML]{EFEFEF} \\
\mlSPF & 56.0 & 0.0010* & {\color[HTML]{B42419} 0.2667} & 0.0240* & {\color[HTML]{B42419} 0.0649} & {\color[HTML]{B42419} 0.1518}\\\hline
\end{tabular}
\caption{Statistical significance between the models in the zero-shot setting. }
\label{table:stat_sig_zero-shot}
\end{subtable}

\begin{subtable}{\textwidth}
\centering
\begin{tabular}{|l|c|lllllll|}
\cline{3-9}
\multicolumn{2}{c|}{}& \lexDict& \lexRand& \LexPred& \ecRand& \ecSPF& \mlRand& \mlSPF\\\cline{2-9}
\multicolumn{1}{c|}{}& \multicolumn{1}{c|}{chrF++} & \multicolumn{1}{c}{55.9} & \multicolumn{1}{c}{57.5}& \multicolumn{1}{c}{58.0}& \multicolumn{1}{c}{56.9} & \multicolumn{1}{c}{57.3} & \multicolumn{1}{c}{57.6} & \multicolumn{1}{c|}{57.5}\\\hline
\lexDict  & 55.9 & \cellcolor[HTML]{EFEFEF} & \cellcolor[HTML]{EFEFEF}& \cellcolor[HTML]{EFEFEF} & \cellcolor[HTML]{EFEFEF} & \cellcolor[HTML]{EFEFEF}& \cellcolor[HTML]{EFEFEF}& \cellcolor[HTML]{EFEFEF} \\
\lexRand &  57.5 &0.0010*& \cellcolor[HTML]{EFEFEF}& \cellcolor[HTML]{EFEFEF} & \cellcolor[HTML]{EFEFEF} & \cellcolor[HTML]{EFEFEF}& \cellcolor[HTML]{EFEFEF}& \cellcolor[HTML]{EFEFEF} \\
\LexPred &  58.0 &0.0010*&  0.0190*& \cellcolor[HTML]{EFEFEF} & \cellcolor[HTML]{EFEFEF} & \cellcolor[HTML]{EFEFEF}& \cellcolor[HTML]{EFEFEF}            & \cellcolor[HTML]{EFEFEF} \\
\ecRand &  56.9 &0.0010*&  0.0040*&  0.0010*& \cellcolor[HTML]{EFEFEF} & \cellcolor[HTML]{EFEFEF}& \cellcolor[HTML]{EFEFEF}& \cellcolor[HTML]{EFEFEF} \\
\ecSPF &  57.3 &0.0010*& {\color[HTML]{B42419}  0.1359} &  0.0030*&  0.0280*& \cellcolor[HTML]{EFEFEF}& \cellcolor[HTML]{EFEFEF}& \cellcolor[HTML]{EFEFEF} \\
\mlRand &  57.6 &0.0010*& {\color[HTML]{B42419}  0.2957} &  0.0310*&  0.0020*& {\color[HTML]{B42419}  0.1149} & \cellcolor[HTML]{EFEFEF}& \cellcolor[HTML]{EFEFEF} \\
\mlSPF&  57.5 &0.0010*& {\color[HTML]{B42419}  0.4096} &  0.0240*&  0.0030*& {\color[HTML]{B42419}  0.1239} & {\color[HTML]{B42419}  0.3137} & \cellcolor[HTML]{EFEFEF} \\
\BT &  58.6 &0.0010* &  0.0010* &  0.0040*&  0.0010*&  0.0010*&  0.0010*&  0.0010*\\\hline
\end{tabular}
\caption{Statistical significance between the models in the non-zero-shot setting.}
\label{table:stat_sig_non-zero-shot}
\end{subtable}
\caption{Statistical significance between models in the zero- and non-zero-shot settings calculated on the chrF++ scores achieved on ArzEn-ST test set CSW sentences. We present the \textit{p}-values and mark \textit{p}-values < 0.05 with $\ast$, where the null hypothesis can be rejected. We include the chrF++ scores for easier readability and comparison.}
\label{table:stat_sig}
\end{table*}

\section{Human Evaluation}
\label{sec:appendix-humanEval}
We present the full results of the human evaluation study discussed in Section \ref{sec:human_eval}. For each evaluated augmentation, we calculate the mean opinion score (MOS) as the average of scores received by the three annotators. In Table \ref{table:MOS_scores}, we present the percentage of sentences falling under each MOS range for understandability and naturalness per augmentation technique. In Table \ref{table:MOS_scores_avg}, we present the average MOS scores per technique.

\begin{table*}[th!]
\centering
\setlength{\tabcolsep}{3pt}
\begin{tabular}{|l | r r r | r r r r | r |}
\hline
\multicolumn{1}{|c|}{MOS}&\multicolumn{1}{c}{\sc Lex$_{Dict}$}&
\multicolumn{1}{c}{\sc Lex$_{Rand}$}&
\multicolumn{1}{c|}{\sc Lex$_{Pred}$}&
\multicolumn{1}{c}{\sc EC$_{rand}$}&
\multicolumn{1}{c}{\sc EC$_{spf}$}&
\multicolumn{1}{c}{\sc ML$_{rand}$}&
\multicolumn{1}{c|}{\sc ML$_{spf}$}&
\multicolumn{1}{c|}{\sc BT}\\\hline
\multicolumn{9}{|c|}{Understandability}\\\hline
1$\leq$*$\<2$  & 35.3 & 4.0& 4.0 & 7.3 & 8.0 & 8.7 & 9.3 & 6.0\\
2$\leq$*$\<3$  & 64.7 & 96.0 & 96.0 & 92.7 & 92.0 & 91.3 & 90.7 & 94.0\\\hline
\multicolumn{9}{|c|}{Naturalness}\\\hline
1$\leq$*$\<2$ & 62.7 & 27.3 & 13.3 & 28.7 & 24.7 & 20.0 & 20.0 & 6.7 \\
2$\leq$*$\<3$ & 21.3 & 25.3 & 20.0 & 22.7 & 25.3 & 19.3 & 25.3 & 13.3 \\
3$\leq$*$\<4$ & 12.0& 21.3 & 27.3 & 27.3 & 27.3 & 32.7 & 31.3 & 26.7\\
4$\leq$*$\leq5$ & 4.0 & 26.0 & 39.3 & 21.3 & 22.7 & 28.0 & 23.3 & 53.3\\\hline
\end{tabular}
\caption{The percentage of synthetic sentences per augmentation technique falling under each mean opinion score (MOS) range for understandability and naturalness, as obtained through human evaluation.
}
\label{table:MOS_scores}
\end{table*}

\begin{table*}[th!]
\centering
\setlength{\tabcolsep}{3pt}
\begin{tabular}{|l | c c c | c c c c | c |}
\cline{2-9}
\multicolumn{1}{c|}{}&\multicolumn{1}{c}{\sc Lex$_{Dict}$}&
\multicolumn{1}{c}{\sc Lex$_{Rand}$}&
\multicolumn{1}{c|}{\sc Lex$_{Pred}$}&
\multicolumn{1}{c}{\sc EC$_{rand}$}&
\multicolumn{1}{c}{\sc EC$_{spf}$}&
\multicolumn{1}{c}{\sc ML$_{rand}$}&
\multicolumn{1}{c|}{\sc ML$_{spf}$}&
\multicolumn{1}{c|}{\sc BT}\\\hline
Understandability & 2.16 & 2.78 & 2.77 & 2.72 & 2.68 & 2.73 & 2.70 & 2.75 \\
Naturalness & 1.80 & 2.84 & 3.34 & 2.74 & 2.84 & 3.08 & 2.96 & 3.76 \\\hline
\end{tabular}
\caption{The average mean opinion scores (MOS) for understandability and naturalness per technique, as obtained through human evaluation.}
\label{table:MOS_scores_avg}
\end{table*}

\end{document}